\documentclass[a4paper,10pt,twoside]{article}

\usepackage{clin}        
\usepackage{harvard}     
\usepackage{lingmacros}
\usepackage{booktabs}
\usepackage{hyperref}

\begin{document}

\title{Opinion aspect extraction in Dutch
children`s diary entries}

\author{Hella Haanstra$^*$ \email{hellahaanstra@gmail.com}\\
{\normalsize \bf Maaike de Boer}$^{**}$ \email{maaike.deboer@tno.nl}\\
\AND \addr{$^*$The Netherlands}
\AND \addr{$^{**}$ TNO, Anna van Buerenplein 1, The Netherlands} }

\maketitle\thispagestyle{empty} 


\begin{abstract}
Aspect extraction can be used in dialogue systems to understand the topic of opinionated text. Expressing an empathetic reaction to an opinion can strengthen the bond between a human and, for example, a robot. The aim of this study is three-fold: 1. create a new annotated dataset for both aspect extraction and opinion words for Dutch children's language, 2. acquire aspect extraction results for this task and 3. improve current results for aspect extraction in Dutch reviews. This was done by training a deep learning Gated Recurrent Unit (GRU) model, originally developed for an English review dataset, on Dutch restaurant review data to classify both opinion words and their respective aspects. We obtained state-of-the-art performance on the Dutch restaurant review dataset. Additionally, we acquired aspect extraction results for the Dutch children’s dataset. Since the model was trained on standardised language, these results are quite promising. \end{abstract}

\section{Introduction}
Emotions play an important role in everyday interactions \cite{sorjonen2012emotion}. Therefore, a better understanding of both the expression of emotion (i.e. opinion) and the underlying emotion in activities and human outings may yield smoother human-computer interaction \cite{brave2003emotion}. Dialogue systems rely on the ability of the computer to recognise what the user is talking about. For example, a computer recognising frustration in a human customer may adapt their response accordingly. In such a response, two things would be important: the polarity of the expression, i.e. whether it is a positive, negative or neutral outing, and the subject of the expression, i.e. what is the customer positive or negative about. These tasks are both part of the field of sentiment analysis. While sentiment analysis covers a broad range of research, most of it is focused on the classification of sentiment polarity. Recognising the subject of an opinion is a more challenging task, known as a specific subtask of Aspect Based Sentiment Analysis (ABSA). This subtask is known as \textit{aspect extraction}. Here, an \textit{aspect} is the subject of the opinion.

Although aspect extraction is broadly researched, almost all research focuses on reviews (e.g. \cite{schouten2016survey}), news articles or Twitter posts. Hence, all data sets and approaches are trained on and aimed at text written by adults. Moreover, most of these studies are focused on English text. 

Aspect extraction for Dutch was introduced in 2016 by the Semantic Evaluation (SemEval) challenge \cite{pontiki2016semeval}, where they provided the first annotated dataset for restaurant reviews in Dutch. In this challenge, only two teams performed in the specific aspect extraction task and their results failed to compete with the English models. The question arises whether the teams competing in the Dutch dataset could not reach state-of-the-art performance due to the architecture of their models, or because Dutch is a more challenging language to extract aspects from than English. 

Dialogue systems specifically aimed at children are less common than dialogue systems aimed at adults. However, research shows that children may benefit from a robot-child relationship in certain situations, such as during hospital visits or at home. One such research has been carried out in cooperation with the Personal Assistant for a healthy Lifestyle (PAL) project\footnote{http://www.pal4u.eu/}, which aims to develop a robot companion for children with diabetes type 1 to assist with self-management and to increase knowledge on coping with diabetes in certain situations (e.g. what to bring on a sleepover). The PAL-project is an experiment which builds on the results of the ALIZ-E project, which showed the potential of a robot-child relationship as an educational tool \cite{coninx2016towards}. Robot companions offer a unique combination  of both a peer and companion as a supporting mediator in difficult situations \cite{baroni2014robotic}. In the PAL-project, the robot makes a physical appearance during hospital visits, and acts as a digital companion at home. Here, the children are encouraged to play games, quizzes and input diary entries about their well-being. For the robot to interact effectively and bond with the child, it must be capable of adaptive social interaction \cite{kanda2004interactive}. Research shows that the quality of a robot-child interaction relies on the robot's ability to adapt its behaviour to the child \cite{belpaeme2012multimodal}. One way to do so is to adequately react to a child's diary entry. For example, if the child writes that it has not been feeling well at school today, the robot could comfort the child or inquire why they were not feeling well. In order to lay a foundation for this, we applied current aspect extraction research to the new domain of Dutch children's text.

The goal of our research was threefold: to create a new annotated dataset for both aspect extraction and opinion word recognition for Dutch children’s language, to acquire aspect extraction results for this children’s dataset and to improve current results for aspect extraction in Dutch reviews. In order to do so, we made use of a deep learning GRU model that has originally been developed for English aspect extraction. First, we will explain the model that was used and the datasets the model has been applied on, namely a Dutch review dataset and a Dutch children's text dataset. Additionally, we tested different sets of word embeddings in order to raise model performance. We found that this model outperformed all other Dutch models in the SemEval aspect extraction task and established a baseline performance for the Dutch children's text model. 

\section{Related work}

Sentiment analysis is executed on three levels: document level, sentence level or aspect level \cite{liu2012sentiment}. Document-level sentiment analysis is considered the simplest of the three, because any supervised learning algorithm can be applied directly since it can be seen as a traditional text classification problem \cite{liu2015sentiment}. A clear disadvantage of this method is that it assumes that a whole document expresses one clear sentiment about a topic. In reality, documents can cover more than one entity and opinions expressed can vary per entity \cite{cambria2017affective}. The aspect level (also known as feature-based opinion mining \cite{yadollahi2017current}), tries to find sentiment-aspect pairs in which sentiment is expressed about a specific aspect. For example, the sentence \textit{'the park was nice but the weather was horrible'} contains two aspects, namely \textit{park} and \textit{weather} and their respective positive and negative sentiments. This aspect based sentiment analysis is especially useful for commercial purposes, in which businesses can focus on specific aspects (e.g. \textit{battery life}) to estimate the public's opinion on them. 

There are many ways to tackle sentiment analysis problems, which results in a large number of different methods being used. These methods have been categorised by various surveys in order to make a more clear distinction. Rajalakshmi et al. \cite{rajalakshmi2017comprehensive} distinguish lexicon based and machine learning approaches for sentiment analysis in general. For aspect analysis and extraction in particular, frequency based methods are a common approach. 



%

Syntactic relations are often used to extract aspects in sentences. Most methods analyse these syntactic relations during preprocessing by running a computational parser. The drawback of using dependency parsers is that they have to be run beforehand and require some feature engineering to perform well \cite{wang2017coupled}. Additionally, grammar and syntactic errors in the preprocessed text may cause the dependency parser to be imprecise. To circumvent these problems, Wang et al. proposed a deep learning method that models the relations between words automatically, without the need of a dependency parser. Machine learning methods are said to be very flexible, but do require a lot of training data and time before being useful \cite{schouten2016survey}. The authors of the 2017 Semantic Evaluation challenge also note that deep learning approaches continue to be popular in sentiment analysis and are often employed by state-of-the-art approaches \cite{rosenthal2017semeval}. We use the model by Wang et al. on two new datasets: a Dutch review dataset and a Dutch children’s dataset.

\section{Method}

\subsection{The Coupled Multi-Layer Attentions model}

The Coupled Multi-Layer Attentions (CMLA) model by Wang et al. extracts aspect and opinion words by making use of attention layers. The attention layers are used to identify the measure of confidence of a word being an aspect (or an opinion word), one attention layer for aspects and opinion words each. The words are classified in one of five categories, based on the BIO encoding scheme. Thus, the final prediction of the model is two vectors $\mathbf{L}^a$ and $\mathbf{L}^p$ which denote the label vectors for aspects and opinion words, respectively. 

\begin{figure}[h]
\centering
	\includegraphics[scale=0.2]{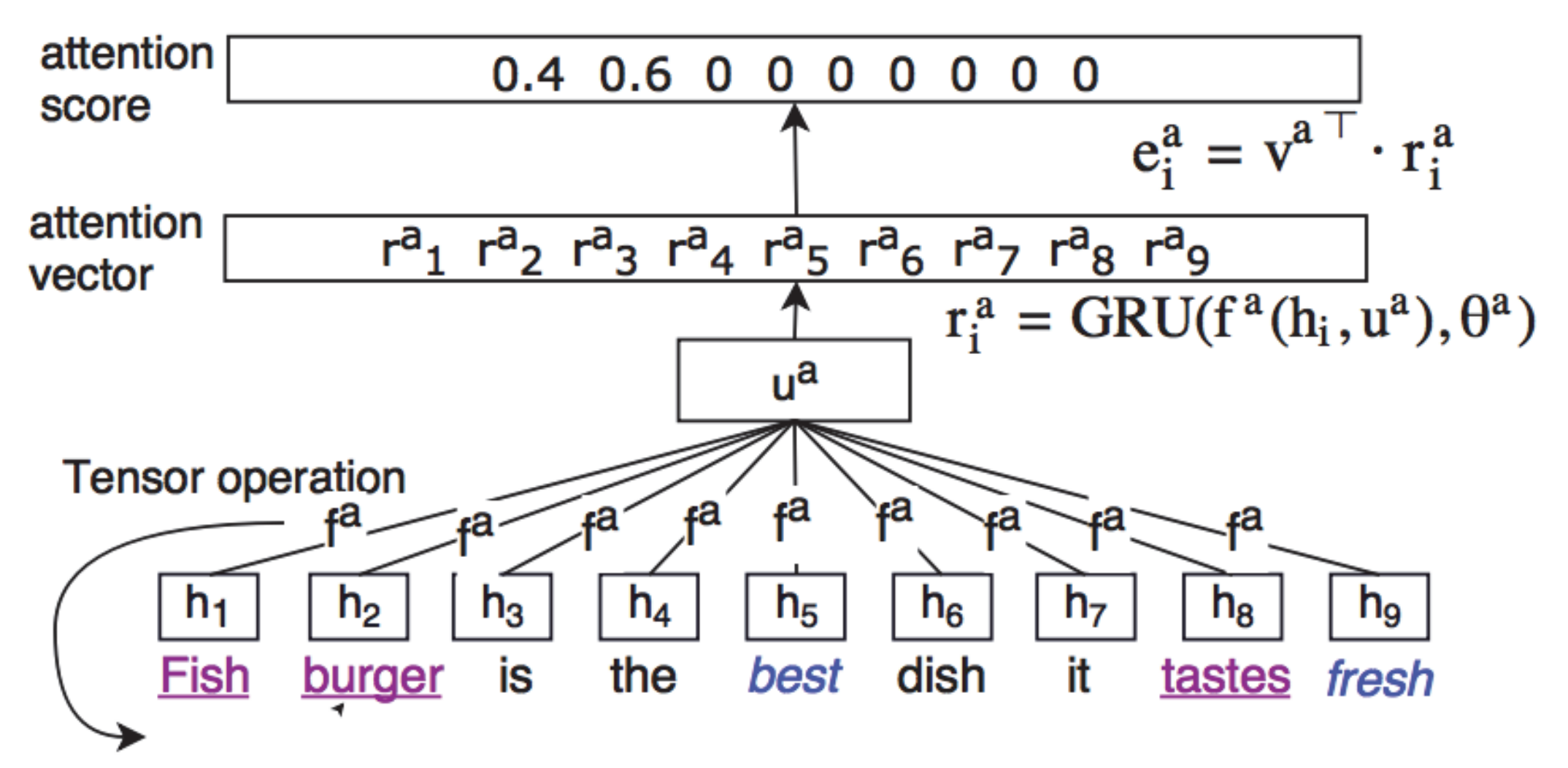}
	\caption{CMLA model layout of one model layer ~\protect\cite{wang2017coupled}}
	\label{fig:wang2017coupleda}
\end{figure}

The layout of the CMLA model is presented in Figure \ref{fig:wang2017coupleda}, taken from the original paper. Depicted is one single layer of the attention model, for aspect extraction. The full model incorporates two layers for each attended classification, which are then coupled. For simplicity, we write $u^a$ and $r^a$ (as is done in Figure \ref{fig:wang2017coupleda}), since both attention models have the same architecture. As we describe the model in general terms, the method applies to both the attention model for aspect extraction (denoted by superscript $a$) as well as the attention model for opinion word extraction (denoted by superscript $p$). Each consecutive part of the model we describe here extends its former part, beginning with the basic attention layer.

Each attention layer tries to learn a \textit{prototype} vector, denoted in Figure \ref{fig:wang2017coupleda} by $u^a$, which will act as a general feature representation for how an aspect (or, in the opinion model, opinion word) would generally look like. Looking at Figure \ref{fig:wang2017coupleda}, the model is executed from the bottom up. 

As a first step, an input sentence is presented to the model, represented by a sequence $h = {h_1, \dots, h_n}$ of trained feature representations. We get these feature representations from pretrained word embeddings \footnote{additional context information is encoded by also applying Gated Recurrent Unit (GRU) to the word embeddings}.Word embeddings are a kind of word representation that can incorporate additional information such as syntactic and semantic properties \cite{turian2010word}. The advantage of this is that it can give text classification networks a head start due to pre-trained information of which words are closely related or alike. For example, the words \textit{king} and \textit{queen} may lie more closely together in vector space than \textit{king} and \textit{apple}. With this knowledge, we include context information even before executing the model. 

For every word feature $h_i$, in the sentence, we then calculate the compositions (i.e. correlations) between every separate word in $h$ and the prototype vector $u^a$, resulting in a composition vector $\beta^a_i \in 	\mathbb{R}^K$ for every word $h_i$. This matrix captures K different kind of compositions between $h_i$ and $u^a$, e.g. a specific syntactic relation. The prototype vector $u$ is first randomly initialised from a uniform distribution $U[-0.2,0.2] \in \mathbb{R}^d$, after which it is continuously updated throughout the training process. 

Next, the model learns a high-level feature vector $r^a$ and an attention score $e^a$ for each word $h_i$ in the sentence. The attention vector $r^a$ is calculated through applying Gated Recurrent Unit (GRU) to the composition vector we acquired from $h$ and $u_a$ in the first step. This GRU is used so that consecutive words may inherit information from their predecessors and context information can be kept.

The Gated Recurrent Unit (GRU) was introduced by Cho et al. \cite{cho2014learning} in 2014 and works as follows. LSTMs are a kind of Recurrent Neural Net, able to maintain a state vector that contains information about past words in a sentence. The GRU works the same as an LSTM unit, but differs in its implementation. Where the LSTM contains four interacting layers, the GRU has three: the update gate, the reset gate and the output gate. The update gate is responsible for how much of the past information (i.e. words) should be passed on to future time steps. The reset gate, on the other hand, decides how much information is to be forgotten. The output gate combines these two with the information of the current time step before passing it on to the next time step. In the example sentence we have seen before, \textit{I highly recommend this beautiful place}, every word may inherit information from its predecessor(s) through the update gate of the GRU, as every word is passed through the GRU in order. 

After applying GRU and retrieving $r^a$, we multiply $r^a$ by a weighting vector $v^a$ to obtain an attention score $e^a$. This attention score defines the likelihood of the word being an aspect. The higher the score, the higher the chance is that the word is an aspect term (or, for the other attention model, opinion word). To obtain a final classification label, a softmax is used on the attention score vector $e^a$, which results in a one-hot vector representation of the classification label.
		
\subsection{Combining the aspect and opinion models}

Since opinion words and aspects are (syntactically) related and thus this relationship could be exploited. Hence, Wang et al.  proposed to \textit{couple} the two attentions to utilise the information contained in both. The two attention vectors are paired, as well as the tensor operator which computes the composition vector $\beta_i$. The new tensor captures the correlations between aspects and opinion words. For example, if \textit{place} has been attended by the aspect attention layer, the correlation relation helps attend \textit{beautiful} as an opinion word. 

\section{Experiments}

\subsection{Datasets}
We made use of three separate datasets: the English restaurant review dataset from SemEval Challenge 2015, the Dutch restaurant review dataset from SemEval Challenge 2016, and the PAL-project dataset containing children's diary entries. All three datasets were annotated following the same SemEval annotation guidelines resulting in an annotation like the below.

\begin{enumsentence}{
\textit{The food was delicious but do not come here
on an empty stomach.} $\rightarrow$
\{category= “FOOD\# QUALITY”, target= “food”,
from: “4”, to: “8”, polarity= “positive”\}}
\end{enumsentence}

Here, \textit{target} is the target of the sentiment expressed in a sentence, i.e. the aspect. \textit{polarity} defines whether the sentiment expressed is positive, neutral or negative. Important to note is the fact that indirect references to entities such as pronouns were annotated as \textit{NULL}. Therefore, the datasets do not contain any indirectly referenced aspects, only explicitly referenced aspects. Additionally, aspects that were not directly related to the entity that was being reviewed (such as other, comparable, restaurants) were considered out of scope and hence not tagged. In this model, all sentences not containing an aspect were labelled \textit{NULL} as well. The SemEval challenges provided a total number of 1,315 and 1,700 training sentences, 685 and 575 test sentences for English and Dutch, respectively. For the children's dataset we manually tagged aspects and tagged opinion words based on a fixed word list for 228 sentences. Since all 228 Dutch sentences were written by children, we preprocessed the sentences by correcting all spelling mistakes manually. The aspect labelling was done by three separate annotators that reached an inter-annotator agreement of 0.60-0.71 between pairs. 

\subsection{Word embeddings}

The CMLA model makes use of pretrained word embeddings as input. The choice of word embedding may therefore influence the performance of the model to an extent. For this reason, we compared a range of different word embeddings in order to maximise performance of the model. All models were obtained by applying word2vec, using \texttt{gensim} \cite{rehurek_lrec}. The first word2vec model was trained on the 2015 Yelp Challenge dataset \footnote{http://www.yelp.com/dataset} by Wang et al. \cite{wang2017coupled}, which we used in the training and classification of the English dataset. In addition, we trained six Word2Vec models for Dutch, including five models trained on text written by adults and one model trained on children's texts. The five regular corpora are the Roularta corpus, a corpus of Wikipedia dump, the SoNaR corpus, the COW corpus and a Combined corpus, which is a combination of the first three \cite{tulkens2016evaluating}. The children's corpus (Basiscript) consists of 86,668 essays written by Dutch elementary school children. Essays cover subjects such as made-up stories, activities they did during weekends or objects and animals (e.g. planes, giraffes). We expect that if we use the children's word embeddings as input for the CMLA model, results for the children's dataset increase and for the SemEval review dataset decrease due to the mark-up of children's language.
		
\section{Results}

\subsection{SemEval Dutch dataset}

We compare our top results of the CMLA system that we used with the
top-performing systems in the SemEval Dutch aspect extraction challenge. For comparison, we show the results of two other competing systems in the challenge in Table \ref{tab:seresultsdutch}. We find that the CMLA model that we used outperforms all other models on this task, including the baseline by a large margin. 

\begin{table}[h]
\centering
\caption{Results for the SE-16 Dutch aspect extraction task ~\protect\cite{pontiki2016semeval}, compared with our CMLA results that are highlighted in bold. * indicate unconstrained systems.}
\label{tab:seresultsdutch}
\begin{tabular}{@{}ll@{}}
\toprule
Team          & F-1    \\ \midrule
\textbf{CMLA} & \textbf{62.84}  \\
IIT-T         & 56.99* \\
TGB           & 51.78 \\ \midrule
Baseline      & 50.64  \\ \bottomrule
\end{tabular}
\end{table}

\begin{figure}[h]
\center
\includegraphics[scale=0.4]{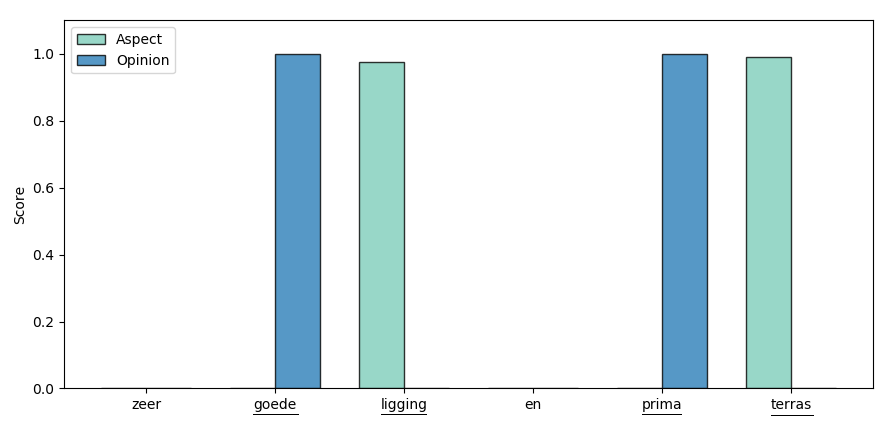}
\caption{Attention scores for a sentence in the SemEval Dutch test set, the top showing correct predictions and the bottom showing an incorrect prediction. Scores are normalised, meaning the higher the score, the more likely it is that the word is an aspect/opinion word. Ground-truth labels are underlined.}
\label{fig:attention_scores_SE-DU}
\end{figure}

We show example output of the model in Figure \ref{fig:attention_scores_SE-DU}. Translation for the sentence depicted in Figure \ref{fig:attention_scores_SE-DU} is given below.

\enumsentence{
		{\textit{zeer goede ligging en prima terras}}\\
		{very good location and great terrace}
		}

\subsection{Children's dataset}

\begin{figure}[h]
\center
\includegraphics[scale=0.4]{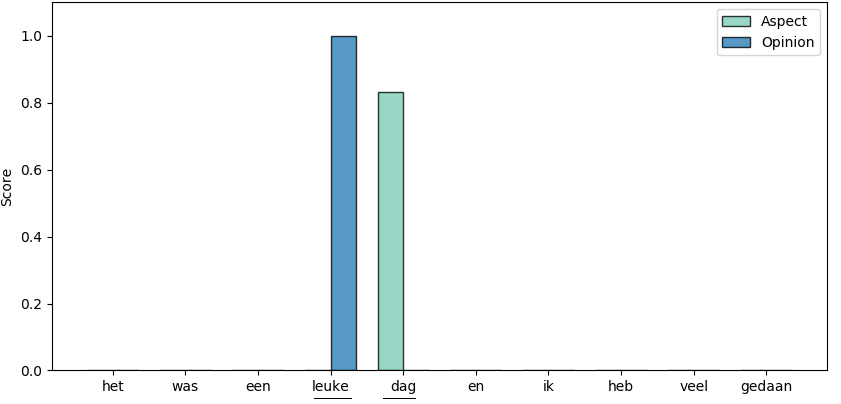}
\caption{Attention scores for a sentence in the PAL test set.}
\label{fig:attention_scores_SE-PAL}
\end{figure}

In addition, we obtained precision and recall results for the Dutch children's dataset, as depicted in Table \ref{tab:precisionrecalldutch}. Figure \ref{fig:attention_scores_SE-PAL} shows an example sentence classified by the model. The translation in English is given below.

\enumsentence{
		{\textit{het was een leuke dag en ik heb veel gedaan}}\\
		{it was a nice day and I did a lot}
		}

\begin{table}[h]
\centering
\caption{Precision and recall scores for both SemEval Dutch and PAL children's dataset. Precision and recall scores are both high for the SemEval Dutch dataset, and we obtain decent precision scores for the PAL children's dataset. Opinon word extraction scores are high for both datasets. }
\label{tab:precisionrecalldutch}
\begin{tabular}{@{}lcccc@{}}
\toprule
         & \multicolumn{2}{c}{aspect} & \multicolumn{2}{c}{opinion} \\ \midrule
         & precision   & recall   & precision   & recall   \\
SemEval Dutch & 55.96       & 71.66    & 96.51       & 92.87    \\
Children's dataset      & 50.00       & 15.56     & 98.89       & 92.07    \\ \bottomrule
\end{tabular}
\end{table}

\section{Discussion and conclusions}

We used a state-of-the-art implementation for English to extract aspects from Dutch text. We see that aspect extraction can be used in dialogue systems to understand the topic of opinionated text. Namely, research shows that the quality of dialogue systems relies on the ability of the computer to recognise what the user is talking about. Ultimately, we aimed to create a children's text dataset with aspect annotations, as well as apply an aspect extraction model for the first time, to lay the foundations of a dialogue system for children.  

For our contributions, we applied the CMLA model on two Dutch datasets, the SemEval Dutch dataset, for which aspect annotations were provided by the challenge, and one that consists Dutch children's text and was collected for the PAL project. We manually annotated the children's dataset for aspect extraction, following the annotation format that is used in the SemEval challenges. In addition, we annotated opinion words for both datasets with the help of the sentiment lexicon by the Pattern module. Our results for the Dutch SemEval dataset outperformed the state-of-the-art models that participated in the challenge. Moreover, the model that we used requires a lot less preprocessing than those models and thus is much easier to use for future datasets. Additionally, we acquired aspect extraction results for the children's set. 

Precision scores for the SemEval Dutch set are significantly lower than recall scores, meaning that the model sometimes extracts aspects that were not labelled as such. It does, however, manage to extract most of the aspects that are known in the dataset, as is shown by the high recall scores. If the extracted aspects are to be used in a dialogue system, it is important that the automatically generated reply suits the conversation. Therefore in this case precision might be more important than recall. While it is important to react adequately to a user, a wrongly tagged aspect might yield a worse user experience than failing to detect an aspect and thus leaving the system with either (1) a general response or (2) repeating a question or asking for more information. In general, the system might be better safe than sorry when replying to a user. 

Interestingly, the precision and recall scores for the children's dataset show a different trend. Precision is significantly higher than recall, meaning the model fails to extract aspects more often than not, but does manage to extract the correct ones when doing so. Interestingly, some aspects that are recognised by the system but were not annotated as such, could arguably be an aspect after all. We also found that sentences that came from one specific subset of the dataset heavily influenced the precision and recall scores. This was a dataset that was obtained by combining two input fields into one sentence. After combination, sentence structure was not always intuitive, nor did it represent a natural sentence. Removing this set increases both precision and recall scores, but we initially decided against this because of the number of sentences that would have to be removed (99/228). With this knowledge, the importance of a large annotated dataset of children's data becomes even greater. Ultimately, we would remove the concatenated dataset completely, and replace it with non-combined sentences in future work. 

In order to create a successful dialogue system, the model must first be able to extract aspects successfully. With the implementation of the state-of-the-art CMLA model for Dutch text, we have established a baseline for aspect extraction in children's text. For future work, it would be especially interesting to explore how other known methods perform in relation to our method. More importantly, we made use of restaurant reviews written by adults as training data. While these share some characteristics with children's text, training on annotated children's text would be the most interesting.  Annotating opinion words is another ambition that arose during our research. We made use of an automatic annotation method, whereas manually labelled sentences are preferred. The results showed that due to this automatic annotation the model learns a fixed word list and, for opinion words, is less flexible in its classification. 

Children's language is particularly challenging due to its irregular- and not rarely incorrect- nature. Although the results for the children's dataset were not as promising as those for the SemEval set, we note that since the model was trained on standardised language, the results are actually quite promising. 


\bibliographystyle{clin} 
\bibliography{bibliography}  

\end{document}